\documentclass[a4paper,fleqn]{cas-dc}
\usepackage[square, numbers]{natbib}
\usepackage{graphicx}
\usepackage{subcaption}
\usepackage{amsmath}
\usepackage{pifont}
\usepackage{array} 
\usepackage{booktabs}
\usepackage{multirow}
\usepackage{graphicx}   
\usepackage{booktabs}   
\usepackage{makecell}   
\usepackage{xcolor}     
\usepackage{tabularx} 
\usepackage{microtype}   
\usepackage{hyphenat}    
\usepackage{graphicx}
\usepackage{caption}
\usepackage{float}     
\usepackage{placeins}  

\usepackage{footmisc}

\usepackage{colortbl}
\usepackage[table]{xcolor}
\usepackage{multirow}
\usepackage{makecell}
\usepackage{booktabs}

\definecolor{mygray}{gray}{0.85}

\usepackage{multirow}
\usepackage{booktabs}
\usepackage{amssymb}

\def\tsc#1{\csdef{#1}{\textsc{\lowercase{#1}}\xspace}}
\tsc{WGM}
\tsc{QE}
\tsc{EP}
\tsc{PMS}
\tsc{BEC}
\tsc{DE}

\begin{document}
\let\WriteBookmarks\relax
\def\floatpagepagefraction{1}
\def\textpagefraction{.001}

\shorttitle{Tri-path DINO for Remote Sensing Multi-Class Change Detection}

\shortauthors{K. Zheng et~al.}

\title [mode = title]{Tri-path DINO: Feature Complementary Learning for Remote Sensing Multi-Class Change Detection}                      

\author[1]{Kai Zheng}\fnmark[1]
\ead{kaizheng@zju.edu.cn} 

\author[2]{Hang-Cheng Dong}\fnmark[1]
\ead{hunsen_d@hit.edu.cn}

\author[3]{Zhenkai Wu}
\ead{zhenkaiwu@zju.edu.cn} 

\author[4]{Fupeng Wei}
\ead{weifupeng@ncwu.edu.cn}

\author[5]{Qin Huang}
\ead{qinhuang@zafu.edu.cn} 

\author[3]{Wei Zhang}
\cormark[2]
\ead{zhangwei@zju.edu.cn} 

\affiliation[1]{organization={Zhejiang University College of Computer Science and Technology},
    city={Hangzhou},
    postcode={310027}, 
    country={China}}

\affiliation[2]{organization={Harbin Institute of Technology School of Instrumentation Science and Engineering},
    addressline={Xidazhi 92}, 
    city={Harbin},
    postcode={150001}, 
    country={China}}

\affiliation[3]{organization={Zhejiang University College of Software Technology},
    city={Hangzhou},
    postcode={310027}, 
    country={China}}

\affiliation[4]{organization={North China University of Water Resources and Electric Power School of Information Engineering},
    city={Zhengzhou},
    postcode={450046}, 
    country={China}}

\affiliation[5]{organization={Zhejiang A\&F University college of Chemistry and Materials Engineering},
    city={Hangzhou},
    postcode={311300}, 
    country={China}}

\fntext[1]{These authors contributed equally to this work.}

\cortext[2]{Corresponding author}

\begin{abstract}
In remote sensing imagery, multi-class change detection (MCD) is crucial for fine-grained monitoring, yet it has long been plagued by feature ambiguity in change regions, which obscures subtle changes and hinders fine-grained discrimination under complex scene variations and scarce annotations.
To address this, we propose the Tri-path DINO architecture, which adopts a multi-path decoupled complementary feature learning strategy to separately capture coarse-grained semantic features and fine-grained target-edge features.
Specifically, we employ the DINOv3 pre-trained model as the backbone feature extraction network to learn coarse-grained features. An auxiliary path also adopts a siamese structure, progressively aggregating intermediate features from the siamese encoder to enhance the learning of fine-grained features. Finally, a multi-scale attention mechanism is introduced to augment the decoder network, where parallel convolutions adaptively capture and enhance contextual information under different receptive fields.
The proposed method achieves state-of-the-art performance on Gaza-change, SECOND, and LandsatSCD, improving mIoU/$F_{\mathrm{scd}}$ by 1.01\%/2.93\%, 0.67\%/2.84\%, and 1.22\%/1.51\%, respectively, over the best competing method.
Grad-CAM visualizations further confirm that the main and auxiliary paths naturally focus on coarse-grained semantic changes and fine-grained structural details, respectively. This synergistic complementarity provides a robust and interpretable solution for advanced change detection tasks, offering a basis for rapid and accurate damage assessment.
\end{abstract}


\maketitle


\section{Introduction}

Remote sensing change detection, which aims to identify regions of interest that have changed between bi-temporal satellite images, plays an irreplaceable role in numerous scenarios such as urban environmental monitoring, agricultural surveillance, disaster assessment, and damage evaluation~\cite{review}. However, Binary Change Detection (BCD)~\cite{marchesi2009ica} merely answers “where changes have occurred,” offering limited actionable information that is often insufficient to guide follow-up actions. Semantic Change Detection (SCD)~\cite{ding2024joint}, on the other hand, requires performing full semantic segmentation on both images before comparing them, which is computationally intensive and demands exhaustive annotation of all semantic categories in both temporal phases. For example, in post-disaster damage assessment, knowing only that “changes have occurred” is far from adequate. It is also essential to clarify “what the changes signify” and to complete the assessment rapidly. Therefore, under real-world constraints, a balance must be struck between efficiency and performance. Multi-class change detection (MCD)~\cite{zhu2024review} addresses this need by subdividing changes into specific semantic categories, thereby providing finer-grained information while significantly reducing annotation effort. This makes MCD a suitable solution for time-sensitive tasks such as rapid damage assessment.

Despite its practical potential, the technical implementation of MCD remains challenging. In real-world scenarios, changed areas are typically small, occupying only a fraction of the total image pixels. Furthermore, due to factors such as damage or degradation, the features distinguishing different change categories are often subtle, making accurate classification particularly difficult. Traditional approaches, including deep learning methods based on siamese networks, commonly suffer from performance degradation due to insufficient information. Additionally, while powerful self-supervised foundation models such as DINOv3~\cite{simeoni2025dinov3} are capable of extracting rich, hierarchical features, they have not yet been fully explored or adapted for this specific task. Direct application often fails to explicitly model and integrate the complementary features needed to distinguish fine-grained inter-class differences within changed regions, such as variations in structural damage severity.

To address the challenges inherent in the MCD task, we propose Tri-path DINO, a novel architecture specifically designed for it. The core innovation of Tri-path DINO is a "third-path" complementary learning strategy, which explicitly extracts and synergizes features at different granularities. The first path serves as the main pathway, utilizing the pre-trained DINOv3 backbone to encode robust, coarse-grained semantic representations. Subsequently, multi-scale intermediate features from the main siamese encoder are progressively aggregated. A Transformer architecture is then employed as an auxiliary siamese encoder to extract fine-grained features. We term this secondary siamese encoder the "third path", corresponding to the main siamese network. Finally, we design a multi-level attention decoder to adaptively refine contextual information across different receptive fields, enabling precise localization and classification of changes.

To demonstrate its performance in real-world scenarios, we conducted experiments on two datasets. The first is the Gaza-change dataset, which we constructed and annotated for infrastructure damage assessment in the Gaza region. This dataset captures six categories of changes in civilian infrastructure between 2023 and 2024. The second is the classic SECOND dataset, whose labels were processed to retain only the changed areas, simulating an MCD task. Experimental results confirm that our model not only achieves leading performance but also possesses practical utility in complex real-world situations. Furthermore, Grad-CAM visualizations provide clear and interpretable evidence: the main path focuses on coarse-grained semantic context, while the auxiliary path concentrates on fine-grained structural details. This visual corroboration highlights the synergistic effect of the three-path design and explains the model’s ability to discern subtle change categories.

The main contributions of this paper are summarized as follows:

\begin{enumerate}
\item We propose Tri-path DINO, a complementary feature learning architecture for MCD that addresses the limited ability of directly adapted visual foundation models to jointly exploit coarse-grained semantic information and fine-grained structural information. The DINOv3-based main path extracts robust semantic representations and local change cues, while the complementary Transformer-based third path aggregates intermediate hierarchical features to model long-range dependencies and fine-grained structural differences, enabling more effective discrimination of subtle inter-class changes.
\item We develop an adaptation and feature-fusion strategy for transferring DINOv3 to complex remote sensing scenarios. LoRA and adapter modules are introduced while retaining the pre-trained DINOv3 weights to incorporate task-specific information, and a Multi-Level Hybrid Attention (MLHA) decoder is designed to fuse the complementary CNN- and Transformer-derived features across multiple receptive fields, thereby improving the localization and classification of sparse and subtle changes.
\item We introduce the Gaza-change dataset for infrastructure damage assessment and comprehensively evaluate Tri-path DINO on both Gaza-change and SECOND. Extensive quantitative comparisons, ablation studies, and Grad-CAM visualizations validate the effectiveness of the proposed components and reveal the complementary roles of the different feature paths, demonstrating the applicability of the proposed framework to fine-grained multi-class change detection.
\end{enumerate}

\section{Related Work}
Remote Sensing Change Detection (RSCD) aims to identify changes in the state of surface objects by analyzing images of the same geographical area acquired at different times. Depending on the task objectives, change detection can be mainly divided into three categories: Binary Change Detection (BCD), Multi-class Change Detection (MCD), and Semantic Change Detection (SCD). In recent years, with the advancement of deep learning, CD methods have achieved remarkable breakthroughs in performance. This section will review the research progress in these three categories and the CD methods based on pre-trained visual foundation models.

\subsection{Remote Sensing Change Detection}

BCD aims to produce a binary mask that labels each pixel as “changed” or “unchanged.” Early approaches relied on traditional machine learning and image processing techniques~\cite{marchesi2009ica}. With the deep learning revolution, siamese architectures based on convolutional neural networks have become mainstream. For example, the FC-Siam network~\cite{FCsiam2018,zhan2017change} employs weight-sharing to extract features from bi-temporal images. To enhance the feature extraction capability of convolutional networks, Fang et al. proposed SNUNet~\cite{denseFCsiam}, which adopts dense skip connections in a U-shaped architecture. Furthermore, the inherently local receptive field of CNNs limits their ability to model long-range dependencies and capture global contextual information. To address this issue, researchers have introduced the attention mechanism and Transformer architecture~\cite{attention,dosovitskiy2020vit}. He et al. proposed a channel bias split attention (CBSA) to recover the information in the change region by adding a simple channel mapping~\cite{he2024cbsasnet}. The Transformer architecture establishes dynamic interactions among all elements within an image by utilizing a self-attention mechanism. This enables the model to explicitly model global dependencies and long-range contextual relationships, which is crucial for distinguishing genuine changes from spurious variations across different regions. In~\cite{BIT}, a hybrid architecture combining Transformer and CNN, termed the BIT model, was proposed. This model employs a Transformer structure to process features obtained via convolution. In contrast, ChangeFormer~\cite{changeformer} adopts a pure Transformer framework. Zhang et al. introduced SwinSUNet~\cite{zhang2022swinsunet}, which incorporates the Swin Transformer~\cite{liu2021swin} structure into the BCD task. These studies demonstrate the critical role of advanced network architecture design in feature extraction.

Both MCD~\cite{zhu2024review} and SCD are inherently segmentation tasks, requiring higher-level semantic understanding compared to BCD. Due to the demand for fine-grained semantic annotations, most publicly available datasets are constructed under the BCD paradigm, resulting in relatively limited research and data resources dedicated to MCD and SCD. MCD aims to classify changed pixels into specific semantic categories. SCD, by nature, is a bi-temporal semantic segmentation task that requires generating precise semantic maps for each temporal phase. Owing to their fundamental similarities, the distinction between MCD and SCD is often not strictly emphasized in the literature~\cite{cui2023mtscd}. Early research in SCD, such as~\cite{mou2018learning}, introduced an end-to-end recurrent convolutional neural network (ReCNN) that integrates the spectral-spatial feature extraction capabilities of CNNs with the temporal modeling strengths of RNNs. Long et al. proposed a multi-task learning framework that decomposes the detection of change regions into three sub-tasks: bi-temporal semantic segmentation, change detection, and boundary detection~\cite{long2025bgsnet}. Ding et al. introduced SCanNet, which designs a semantic learning loss based on temporal consistency constraints~\cite{ding2024joint}. Similarly, in~\cite{TANG2024299}, BCD is also incorporated as a subtask, and a contrastive learning module is designed to interact information between semantic detection and change detection.

\subsection{Visual Foundation Models}

Visual Foundation Models (VFMs) acquire powerful general visual representation capabilities through pre-training on large-scale, multi-source datasets. This opens new perspectives for visual change detection. CLIP~\cite{clip} learns from image-text pairs to align visual and textual representations, enabling open-vocabulary reasoning. Segment Anything Model (SAM)~\cite{SAM} demonstrates exceptional zero-shot generalization capability on segmentation tasks. In contrast, the DINO series~\cite{simeoni2025dinov3} relies solely on self-supervised learning to extract general-purpose visual features from massive datasets, making it well-suited as a feature encoder.

The success of VFMs has naturally attracted significant interest from researchers, and how to adapt and inject domain-specific knowledge into these foundation models has become a current focus. A common paradigm is to adopt SAM as the backbone, design it in a siamese architecture, and then train it on the target domain~\cite{qin2025sam2}. Furthermore, fine-tuning these VFMs has led to further performance improvements in change detection. For instance, \cite{gao2025combining} designed CNN adapters for this purpose. Meanwhile, \cite{zhang2024integrating} introduced a parameter-efficient fine-tuning approach using LoRA-like adapters. To mitigate the substantial parameter overhead of VFMs, \cite{ding2024adapting} alternatively employed lightweight pre-trained variants such as FastSAM~\cite{fastsam} and MobileSAM~\cite{zhang2023faster}. 

Despite the powerful representational capabilities of models like SAM, their robustness on remote sensing images remains limited, often struggling to identify RS-specific targets comprehensively. This significantly hinders the application of VFMs in SCD and MCD tasks. \cite{mei2024scd} first introduced SAM into the SCD task, meticulously designing a dual encoder-decoder architecture. This approach cross-fuses features from MobileSAM and a CNN encoder, and employs a multi-task scheme in the decoder to progressively decode features at different levels. \cite{jiang2025adaptvfms} proposed a large-scale remote sensing dataset for fine-tuning and simultaneously utilized both CLIP and SAM to encode bi-temporal features. \cite{gaza} was the first to incorporate DINO as a feature extractor for MCD, also adopting a multi-scale feature fusion architecture to enhance detection performance. In summary, effectively leveraging the potential of VFMs for high-level semantic tasks like SCD and MCD necessitates deliberate design of data and models.

\section{Method}
In this section, we detail the design and training specifics of the proposed Tri-path DINO, covering the backbone siamese network architecture, the third-path siamese network structure, the decoder network architecture, and the composition of the loss function.
\subsection{Overall}

\begin{figure*}
    \centering
    \includegraphics[width=1.0\linewidth]{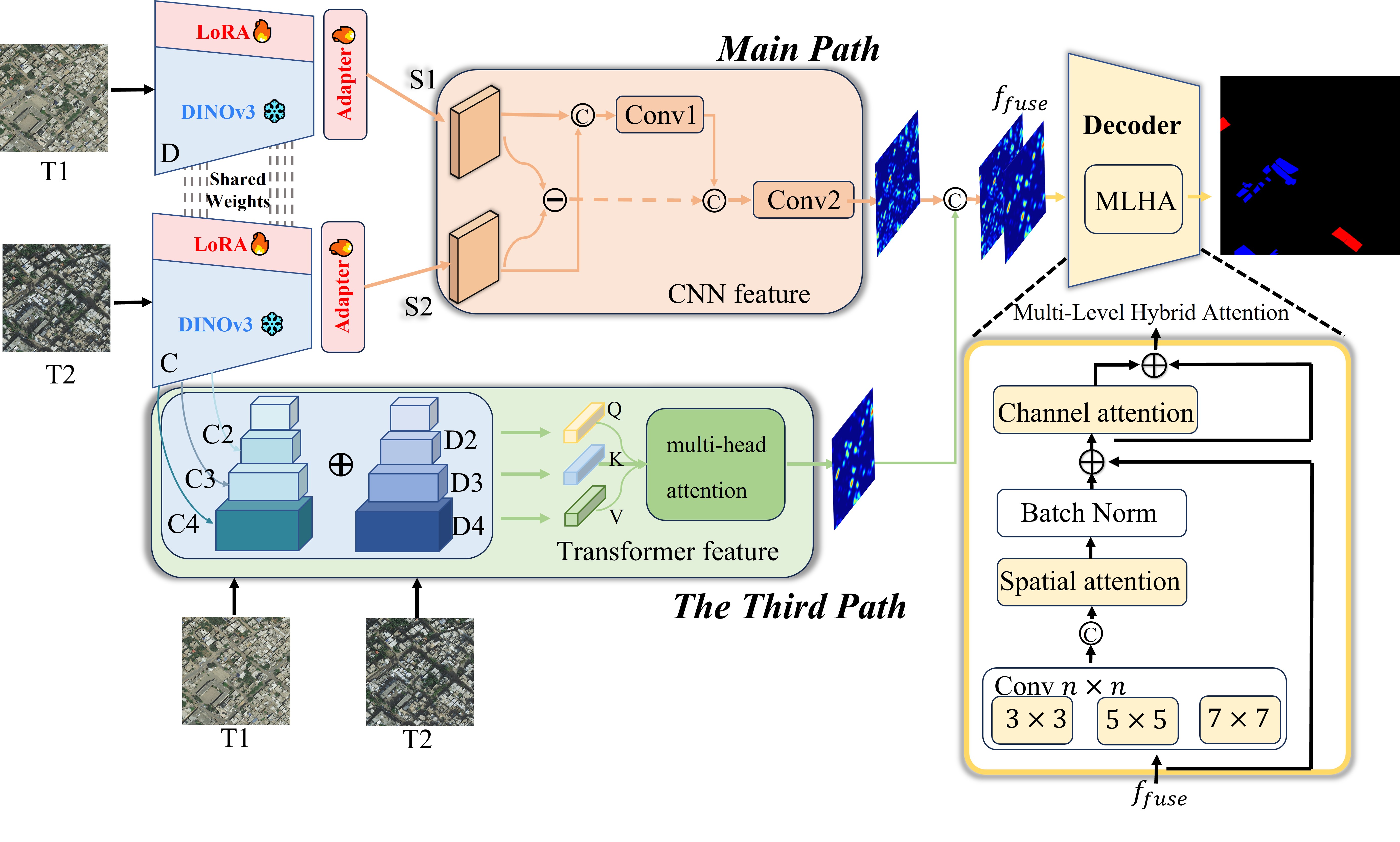}
    \caption{Overall architecture of the proposed Tri-path DINO. Our main path consists of a Siamese-structured convolutional network designed to extract local information, thus favoring more fine-grained details. In contrast, the third path adopts a transformer architecture to capture global information. Owing to this complementary structural design, the main path and the third branch are induced to learn the features they each excel at, namely coarse-grained and fine-grained features.}
    \label{fig:1}
\end{figure*}

The overall architecture of the proposed Tri-path DINO is illustrated in Figure \ref{fig:1}. It follows a siamese encoding and unified decoding paradigm, designed to extract and fuse complementary features from bi-temporal remote sensing images $T_1$ and $T_2$. The framework consists of three cooperative parts, (1)the pre-trained DINOv3 backbone with CNN-based feature extractor, (2) a Transformer feature extractor and (3) decoder. The three paths refer to the two temporal encoding branches with shared weights that process $T_1$ and $T_2$, respectively, and the third pathway that uses their intermediate features for context modeling. The first two branches and the subsequent CNN jointly form the main Siamese pathway. The decoder fuses the main-path and third-path features and is not counted as a separate path.

\subsection{DINO Backbone}
Firstly, we introduce the backbone of Tri-path DINO. The backbone semantic encoder is built upon a pre-trained DINOv3. Given a pair of input images $T_1$ and $T_2$, they are independently processed by the shared DINOv3 backbone, which terms as DINO siamese backbone. Subsequently, while keeping the original pre-trained weights frozen, we insert two types of lightweight adaptation modules, LoRA and adapter, into the backbone. As the primary feature extraction path, let the output features be denoted as $S_1$ and $S_2$, respectively. Following this, a CNN feature extractor is appended. Denoting the output of the CNN feature extractor as $f_{CNN}$, we obtain:
\begin{equation}
    f_{CNN}=Conv_2([S_1-S_2,Conv_1([S_1,S_2])]),
\end{equation}
where $Conv_1$ and $Conv_2$ are two convolutional feature processors, each consisting of a $3\times3$ convolution followed by batch normalization, a GELU activation function, and a $1\times1$ convolution.

Through multi-level extraction of the backbone features $S_1$ and $S_2$, and by employing a carefully designed CNN feature extractor to process these features along with their difference feature, the backbone is guided to focus on local, coarse-grained targets. The shared encoder applies a consistent feature mapping to both temporal inputs, while differencing and concatenation highlight temporal discrepancies and retain bi-temporal semantic context, respectively. Sharing the encoder weights also avoids duplicating the backbone parameters, while both temporal inputs still require forward computation.

\subsection{The Third Path}

To model long-range dependencies while exploiting hierarchical representations from different depths of the DINO backbone, we employ a cross-layer multi-head attention mechanism. Let
\[
X_2=C_2+D_2,\qquad X_3=C_3+D_3,\qquad X_4=C_4+D_4.
\]
The projected query, key, and value are defined as
\[
Q=X_2W_Q,\qquad K=X_3W_K,\qquad V=X_4W_V,
\]
and the cross-layer attention output is given by
\[
f_{\mathrm{Tr}} = \operatorname{Softmax}\left(\frac{QK^{\top}}{\sqrt{d_k}}\right)V.
\]
Unlike standard self-attention, in which $Q$, $K$, and $V$ are derived from the same representation, the proposed formulation explicitly aggregates information across hierarchical feature depths. Here, $X_2$ provides a relatively structure-sensitive reference for querying, $X_3$ provides an intermediate representation for constructing cross-layer affinities, and $X_4$ provides deeper semantic information to be selectively aggregated. Consequently, the resulting representation can be interpreted as a structure-guided semantic aggregation, in which deeper contextual information is adaptively retrieved according to its compatibility with intermediate structural representations. This design is intended to preserve spatially informative cues while enriching them with higher-level semantics for pixel-level multi-class change detection.

\subsection{Enchaned Decoder}
After completing the feature extraction from the two complementary paths, all features are fed into the decoder module located on the right side of Figure \ref{fig:1} for integration and refinement, ultimately generating a pixel-level multi-class change map. The core innovation of our decoder is the Multi-Level Hybrid Attention (MLHA), which adaptively fuses and enhances features from multi-scale and multi-semantic levels. Specifically, let the concatenated feature obtained from the CNN features and Transformer features be denoted as $f_{fuse}=[f_{CNN},f_{Tr}]$. The MLHA module comprises two stages, initial fusion stage and final fusion stage. At the initial fusion stage, we process the concatenated features using parallel convolutional kernels (with kernel sizes of $3\times3$,$5\times5$, and $7\times7$) to capture contextual patterns at different receptive fields. We have:
\begin{equation}
    F_k = GELU( BN( Conv_k(X) ) ),k=1,2,...,K,
\end{equation}
\begin{equation}
    F_{concat} = Concat(F_1,F_2,…,F_K),
\end{equation}
\begin{equation}
    X_{spatial} = f_{fuse}\odot \sigma( BN( Conv_{1\times1}(F_{concat}) ) ),
\end{equation}
\begin{equation}
     y=f_{fuse} + GELU( BN( Conv_{3\times3}(X_{spatial}) ) ).
\end{equation}
At the final fusion stage, we employ the commonly used channel attention~\cite{hu2018squeeze} mechanisms to further extract distinct types of focused features. Then we have:
\begin{equation}
    out = y + y_{channel}.
\end{equation}

\subsection{Loss Function}

In multi-class change detection for remote sensing imagery, the regions of interest (i.e., changed areas) typically occupy only a small fraction of the image. Moreover, severe class imbalance exists among different change categories. To mitigate these training challenges, various weighted loss functions have been proposed. In this work, we adopt a composite loss combining Focal Loss~\cite{lin2017focal}, Dice Loss~\cite{milletari2016v}, and Lovász Loss~\cite{berman2018lovasz} to jointly address the class-imbalance issue. Focal Loss down-weights well-classified examples and focuses on hard samples; Dice Loss directly optimizes the overlap between predictions and ground truth; and Lovász Loss, as a differentiable surrogate for the Jaccard index, helps refine segmentation boundaries. The combination of these losses ensures stable gradient propagation and encourages the model to achieve balanced performance across all change categories. Then, we have:

\begin{equation}
    \mathcal{L}_{total}= \alpha\mathcal{L}_{focal}+ \beta\mathcal{L}_{Dice}+(1-\alpha-\beta)\mathcal{L}_{LOVASZ}.
\end{equation}

Here, $\alpha$, $\beta$, and $\gamma$ denote the weights of the Focal, Dice, and Lov\'asz loss terms, respectively, with $\gamma=1-\alpha-\beta$, so that $\alpha+\beta+\gamma=1$. The default values are $\alpha=0.4$, $\beta=0.3$, and $\gamma=0.3$.

\section{Experiments}
\subsection{Datasets}
SECOND-CD~\cite{yang2021asymmetric} is a large-scale, high-quality dataset specifically constructed for semantic change detection. The imagery covers three major Chinese cities: Hangzhou, Chengdu, and Shanghai, comprising a total of 4,662 paired high-resolution aerial/satellite images with a size of 512×512 pixels (spatial resolution 0.3–5 m). The dataset is split into 2,968 pairs for training and 1,694 pairs for testing. It provides bi-temporal semantic labels and change semantic labels, encompassing categories such as non-vegetated ground surface, tree, low vegetation, water, buildings, and playgrounds, along with their corresponding change information.

Gaza-change~\cite{gaza} is a dataset for multi-class change detection in the Gaza Strip. It comprises 922 precisely co-registered image pairs, each of 512×512 pixels, captured by the Beijing-2 satellite (0.8 m Ground Sampling Distance) between 2023 and 2024. The imagery covers nine major urban areas, including Khan Yunis and Rafah. Following the annotation paradigm of MCD tasks, where only changes are annotated. The dataset labels the target pixel into one of six semantic change categories: building damage, new building, new camp, farmland damage, greenhouse damage, and new greenhouse. The dataset is randomly split into 554 pairs for training, 184 for validation, and 184 for testing.

LandsatSCD~\cite{landsatscd} is a public dataset for remote sensing semantic change detection. Each image has a spatial size of $416\times416$ pixels. In the present MCD experiments, the evaluation labels comprise one no-change background category and nine change categories. The dataset is split into 1,431 image pairs for training, 477 for validation, and 477 for testing.

\subsection{Implementation Details}

All experiments are implemented in PyTorch and conducted on a single NVIDIA RTX 3090 Ti GPU. \textbf{The experiments on Gaza-change, SECOND, and LandsatSCD use the same training settings:} AdamW with an initial learning rate of $1\times10^{-4}$ and a weight-decay coefficient of 0.05, training for 150 epochs, and validation at the end of each epoch.

Tri-path DINO uses a ConvNeXt-Tiny backbone pretrained with DINOv3 and introduces LoRA and adapters for task adaptation while keeping the original pretrained weights frozen. LoRA is applied to the MLP linear projections, \texttt{pwconv1} and \texttt{pwconv2}, in all 18 ConvNeXt blocks, with rank $r=16$ and a scaling parameter of 32. Adapters use a bottleneck structure and are attached as parallel post-block branches to the outputs of the last four backbone blocks: the final block of stage 3 and all three blocks of stage 4. The corresponding input channel dimensions are 384 and 768, and the bottleneck dimensions are 96 and 192, respectively, giving a bottleneck dimension equal to one-quarter of the input channels.

The MLHA module employs parallel convolution kernels of sizes $3\times3$, $5\times5$, and $7\times7$ to capture local and contextual information at different receptive fields. The composite loss consists of Focal, Dice, and Lov\'asz losses, with weights $\alpha=0.4$, $\beta=0.3$, and $\gamma=0.3$, respectively, following Zheng et al.~[43]. These objectives emphasize difficult samples, prediction--ground-truth overlap, and IoU optimization, respectively.

Furthermore, to adapt the comparative methods to the MCD task, the output categories of these methods were correspondingly modified. It should be noted that the SECOND dataset provides pixel‑wise labels only for changed regions and does not supply separate bi‑temporal semantic labels. We evaluated our method by comparing it with a range of state-of-the-art techniques that represent diverse methodological lines. The selected baselines include: BIT~\cite{BIT}, SNUNet~\cite{fang2021snunet}, LGPNet~\cite{LPGNet}, ChangeFormer~\cite{changeformer}, USSFCNet~\cite{USSFCNet}, DDLNet~\cite{ma2024ddlnet}, Rsmamba~\cite{liu2024rsmamba}, CDxLstm~\cite{wu2025cdxlstm}, AnyChange~\cite{anychange}, and MC-DiSNet~\cite{gaza}.

\subsection{Evaluation Metrics}

To comprehensively evaluate the performance of MCD tasks, we adopt the four metrics proposed in~\cite{ding2022bi} (identical to those used for SCD): Overall Accuracy (OA), mean Intersection over Union (mIoU), Separated Kappa (SeK), and the SCD variant of the F1-score (denoted as $F_{SCD}$). OA is defined as follows:

\begin{equation}
    \mathrm{OA} = \frac{\sum_{i=0}^{N} q_{ii}}{\sum_{i=0}^{N}\sum_{j=0}^{N} q_{ij}},
\end{equation}
where $q_{ij}$ denotes an element of the confusion matrix, representing the number of pixels whose ground-truth class is $j$ but are predicted as class $i$. Here, class 0 corresponds to the background (no change). OA measures the proportion of all pixels that are correctly classified.

In this study, the mean Intersection over Union (mIoU) considers both the ``no-change'' and ``change'' categories by averaging their respective IoU values. Its formula is given as follows:
\begin{equation}
    mIoU= \frac{IoU_{nc}+IoU_{c}}{2},
\end{equation}

\begin{equation}
    IoU_{nc} = \frac{q_{00}}{\sum_{i=1}^{N}q_{i0}+\sum_{j=1}^{N}q_{0j}-q_{00}},
\end{equation}

\begin{equation}
    IoU_{c} = \frac{\sum_{i=1}^{N}\sum_{j=1}^{N}q_{ij}}{\sum_{i=0}^{N}\sum_{j=0}^{N}q_{ij}-q_{00}}.
\end{equation}

The SeK coefficient specifically assesses the model's performance on the critical task of identifying "changed" regions. Let $\hat{q_{ij}}=q_{ij}$ except for entries corresponding to the “no-change” (background) class. Then the formula is:

\begin{equation}
    \rho = \frac{\sum_{i=1}^{N} \hat{q_{ii}}}{\sum_{i=0}^{N}\sum_{j=0}^{N}\hat{q_{ij}}},
\end{equation}
\begin{equation}
    \eta = \frac{\sum_{i=0}^{N} (\sum_{j=0}^{N} \hat{q_{ij}} \cdot \sum_{j=0}^{N} \hat{q_{ji}})}{(\sum_{i=0}^{N}\sum_{j=0}^{N}\hat{q_{ij}})^2},
\end{equation}

\begin{equation}
    Sek = e^{IoU_c-1}\cdot\frac{(\rho-\eta)}{(1-\eta)}.
\end{equation}

Similarly, $F_{scd}$ is computed exclusively within regions annotated as “changed.” The metric first calculates precision and recall considering only the changed-class pixels, and then derives their harmonic mean as follows:
\begin{equation}
    P_{scd} =\frac{ \sum_{i=1}^{N} q_{ii}}{\sum_{i=1}^{N}\sum_{j=0}^{N} q_{ij}},
\end{equation}
\begin{equation}
    R_{scd} =\frac{ \sum_{i=1}^{N} q_{ii}}{\sum_{i=0}^{N}\sum_{j=1}^{N} q_{ij}},
\end{equation}

\begin{equation}
    F_{scd} = \frac{2\cdot P_{scd}\cdot R_{scd}}{P_{scd}+R_{scd}}.
\end{equation}
\subsection{Main Results}

\begin{figure*}[htpb]
    \centering
    \includegraphics[width=1.0\linewidth]{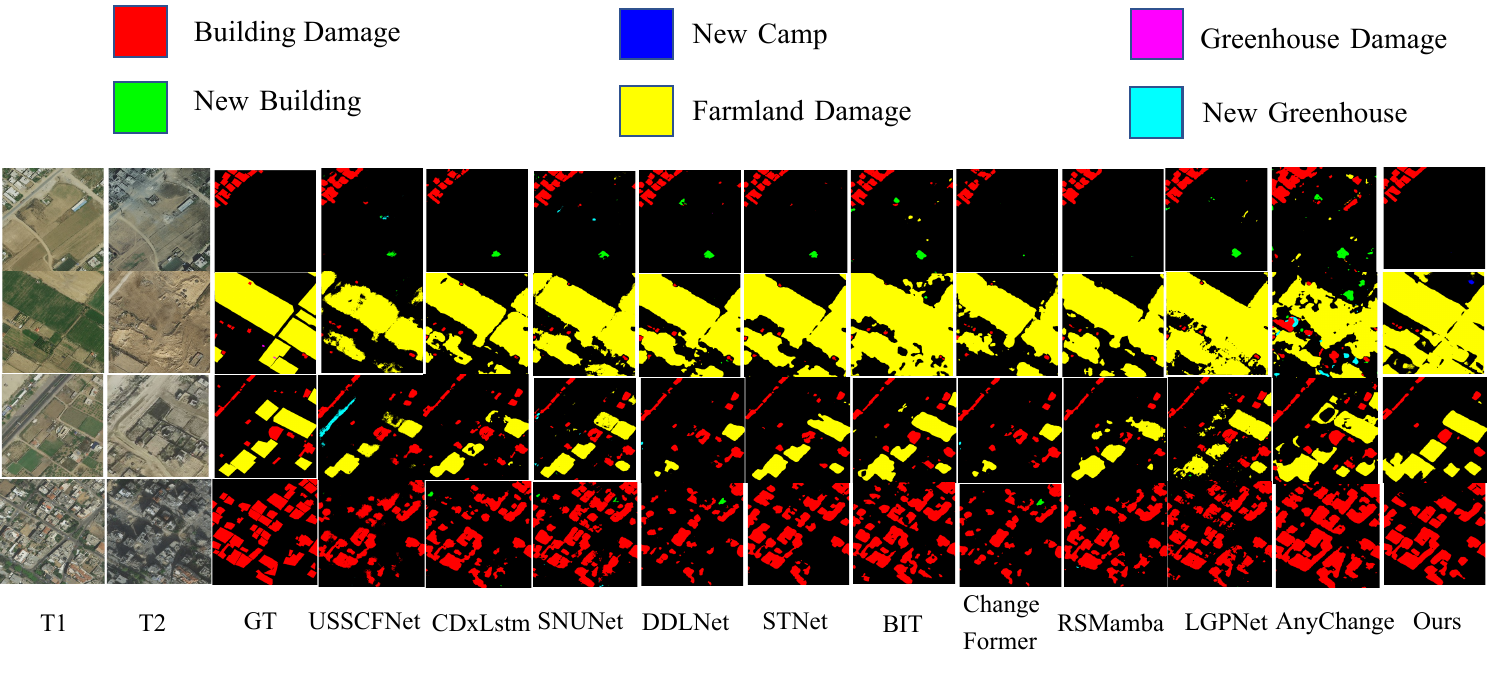}
    \caption{Example results on the Gaza-Change dataset.}
    \label{fig:2}
\end{figure*}

\begin{figure*}[htpb]
    \centering
    \includegraphics[width=1.0\linewidth]{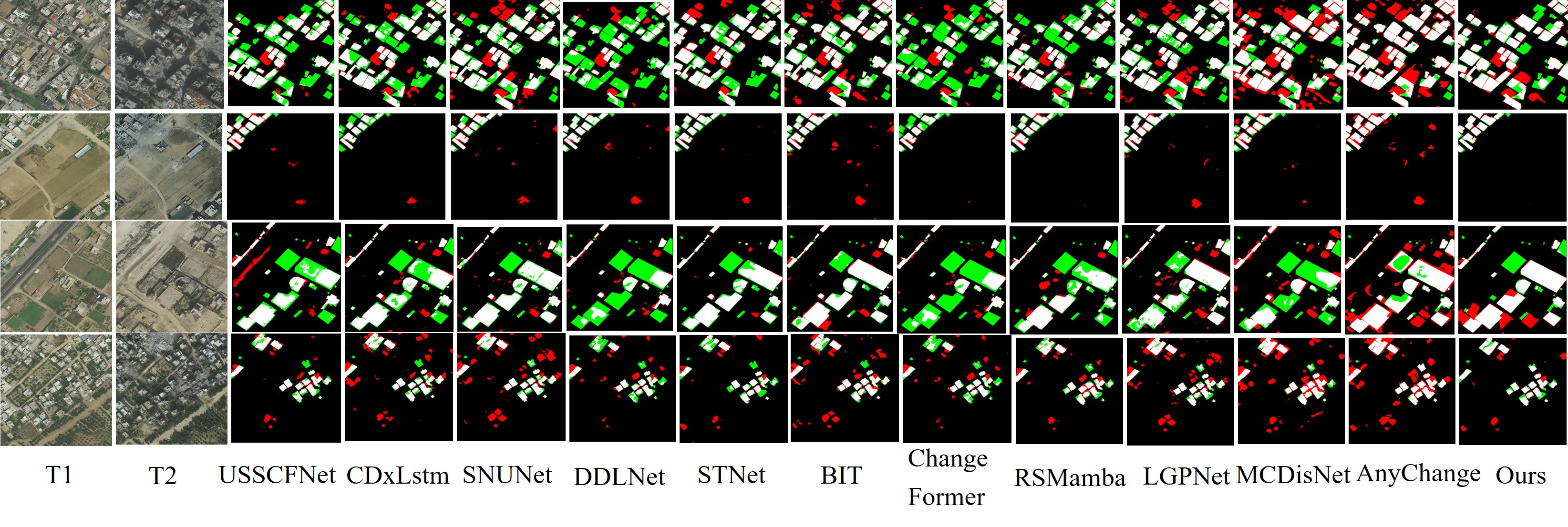}
    \caption{Qualitative comparison on the Gaza-Change dataset, with color coding: true positives (white), true negatives (black), false positives (red), and false negatives (green).}
    \label{fig:3}
\end{figure*}

\begin{figure*}[htpb]
    \centering
    \includegraphics[width=1.0\linewidth]{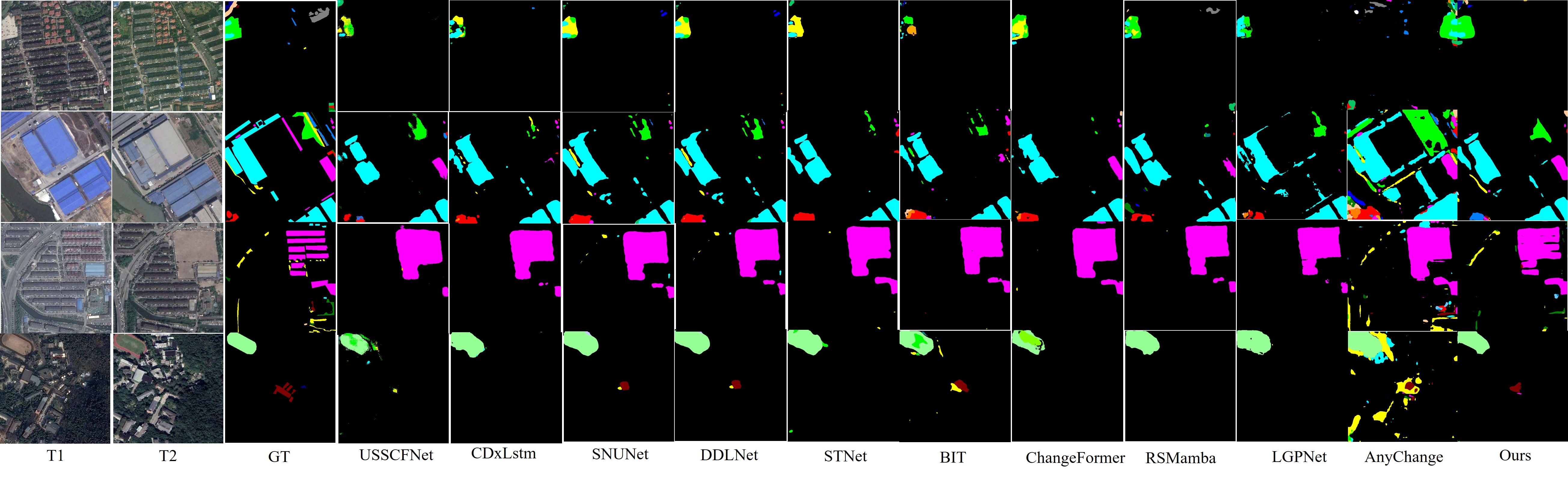}
    \caption{Example results on the SECOND dataset.}
    \label{fig:second1}
\end{figure*}

\begin{figure*}[htpb]
    \centering
    \includegraphics[width=1.0\linewidth]{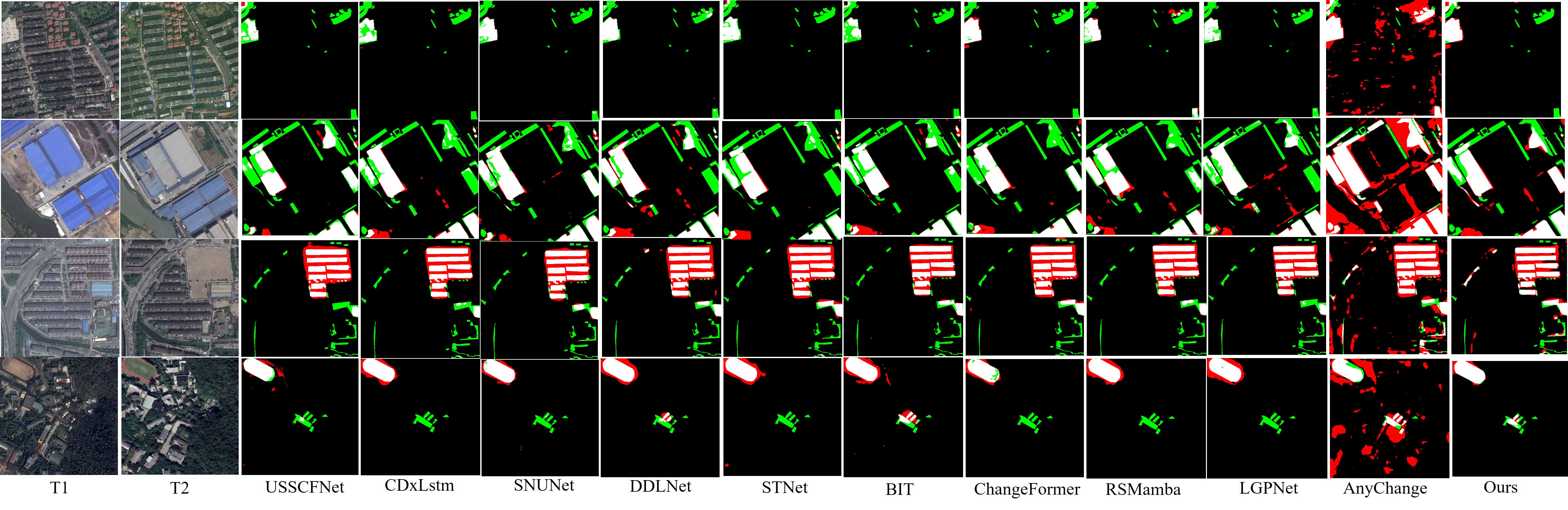}
    \caption{Qualitative comparison on the SECOND dataset, with color coding: true positives (white), true negatives (black), false positives (red), and false negatives (green).}
    \label{fig:second2}
\end{figure*}

\begin{figure*}[htpb]
    \centering
    \includegraphics[width=1.0\linewidth]{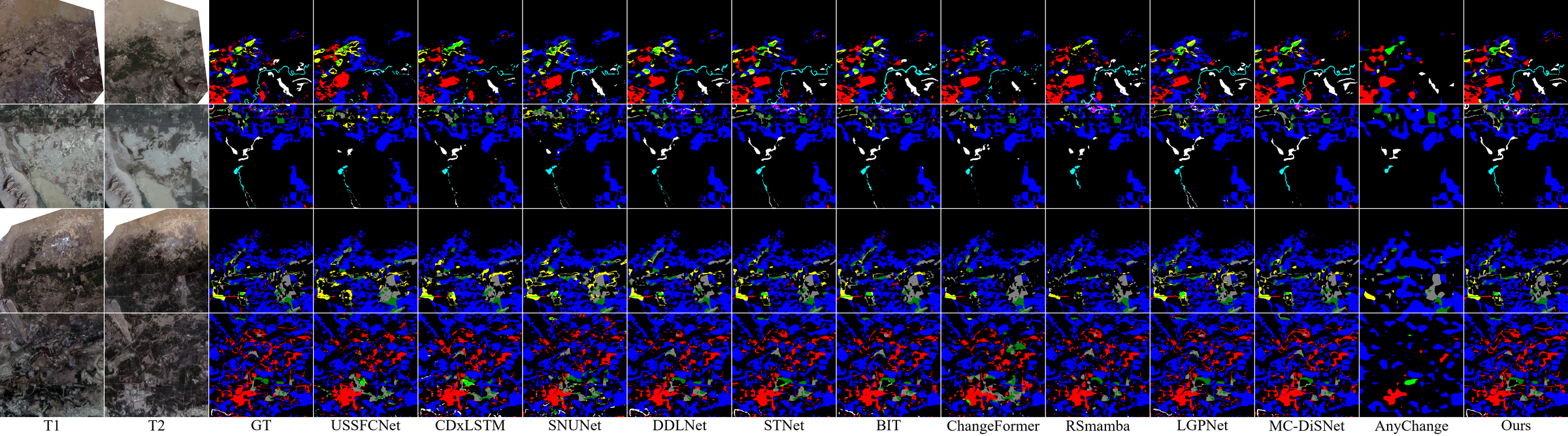}
    \caption{Example results on the LandsatSCD dataset.}
    \label{fig:landsatSCD1}
\end{figure*}

\begin{figure*}[htpb]
    \centering
    \includegraphics[width=1.0\linewidth]{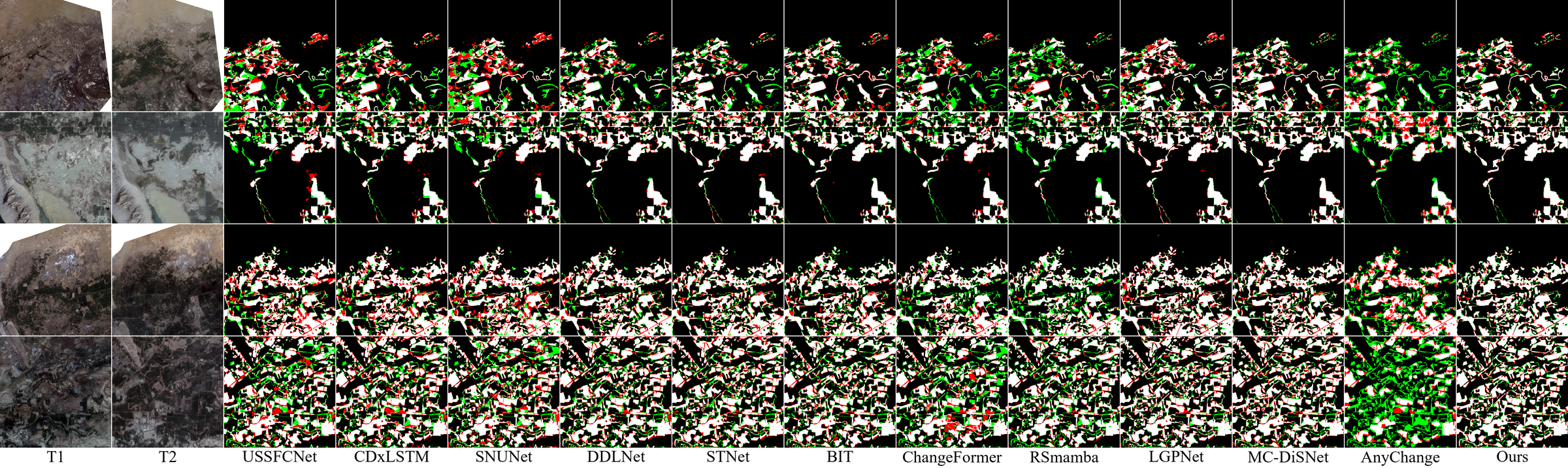}
    \caption{Qualitative comparison on the LandsatSCD dataset, with color coding: true positives (white), true negatives (black), false positives (red), and false negatives (green).}
    \label{fig:landsatSCD2}
\end{figure*}

As shown in Figure~\ref{fig:2} and Figure~\ref{fig:3}, we present a visual comparison of the detection results of different methods on the Gaza-change dataset. Tri-path DINO produces clearer detection boundaries, indicating that the proposed complementary learning architecture enables the model to better understand the detailed features of changes. Meanwhile, by observing the results in the first row of Figure~\ref{fig:2} and the second row of Figure~\ref{fig:3}, it can be seen that the proposed method improves the recognition rate of small targets and avoids the false detections encountered by other methods, demonstrating the high detection performance of Tri-path DINO for fine-grained features. For larger targets, Tri-path achieves the most complete detection. This ability to adapt to targets of varying sizes strongly demonstrates the potential of the complementary learning framework in remote sensing tasks.

Compared with the facility damage, the SECOND dataset contains more detailed variations. Figure~\ref{fig:second1} and Figure~\ref{fig:second2} present the detection results of various state-of-the-art methods on this dataset. It can be clearly observed that the proposed Tri-path DINO captures finer details. In particular, the gaps between buildings in the third row are identified, a capability not achieved by other methods. Additionally, the subtle changes in the central area of the fourth row are also recognized, reflecting the sensitivity of complementary learning to changes. 

Figures~\ref{fig:landsatSCD1} and~\ref{fig:landsatSCD2} present the detection results of different methods on LandsatSCD. The proposed Tri-path DINO captures larger changed regions while preserving fine local details. In particular, the narrow changed structures in the second row are retained, and the fragmented regions in the fourth row show closer agreement with the ground truth. The binary comparisons also show fewer false detections and missed changes than several competing methods. In summary, the complementary learning framework significantly enhances the discriminative ability of the pre-trained model for fine-grained features.

\begin{figure*}[htpb]
    \centering
    \includegraphics[width=0.5
    \linewidth]{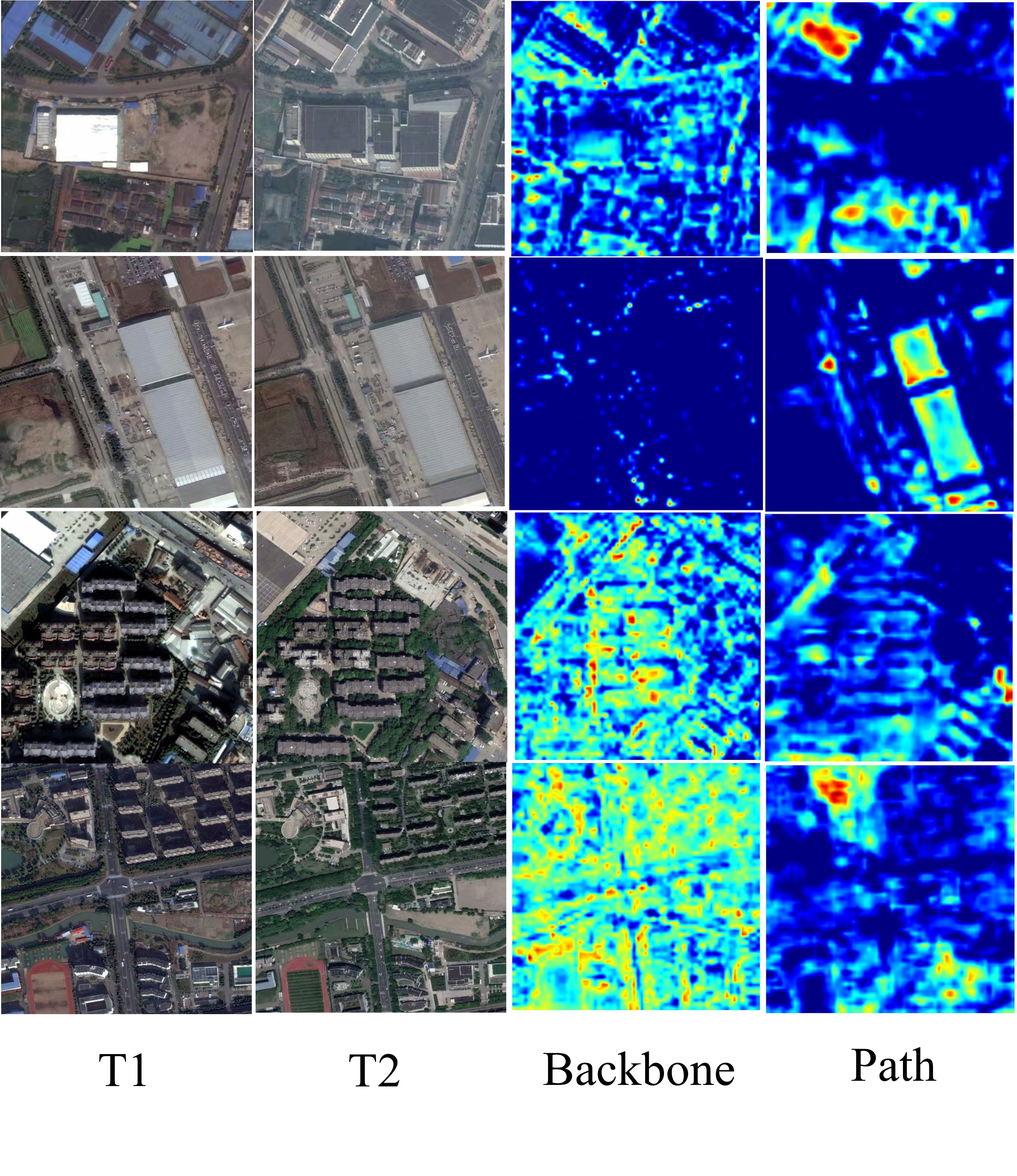}
    \caption{Heatmaps based on the GradCAM visualization of the SECOND dataset. Path represents the features learned by the third pathway, while Backbone represents the features learned by the backbone siamese network.}
    \label{fig:4}
\end{figure*}

\begin{figure*}[htpb]
    \centering
    \includegraphics[width=0.5
    \linewidth]{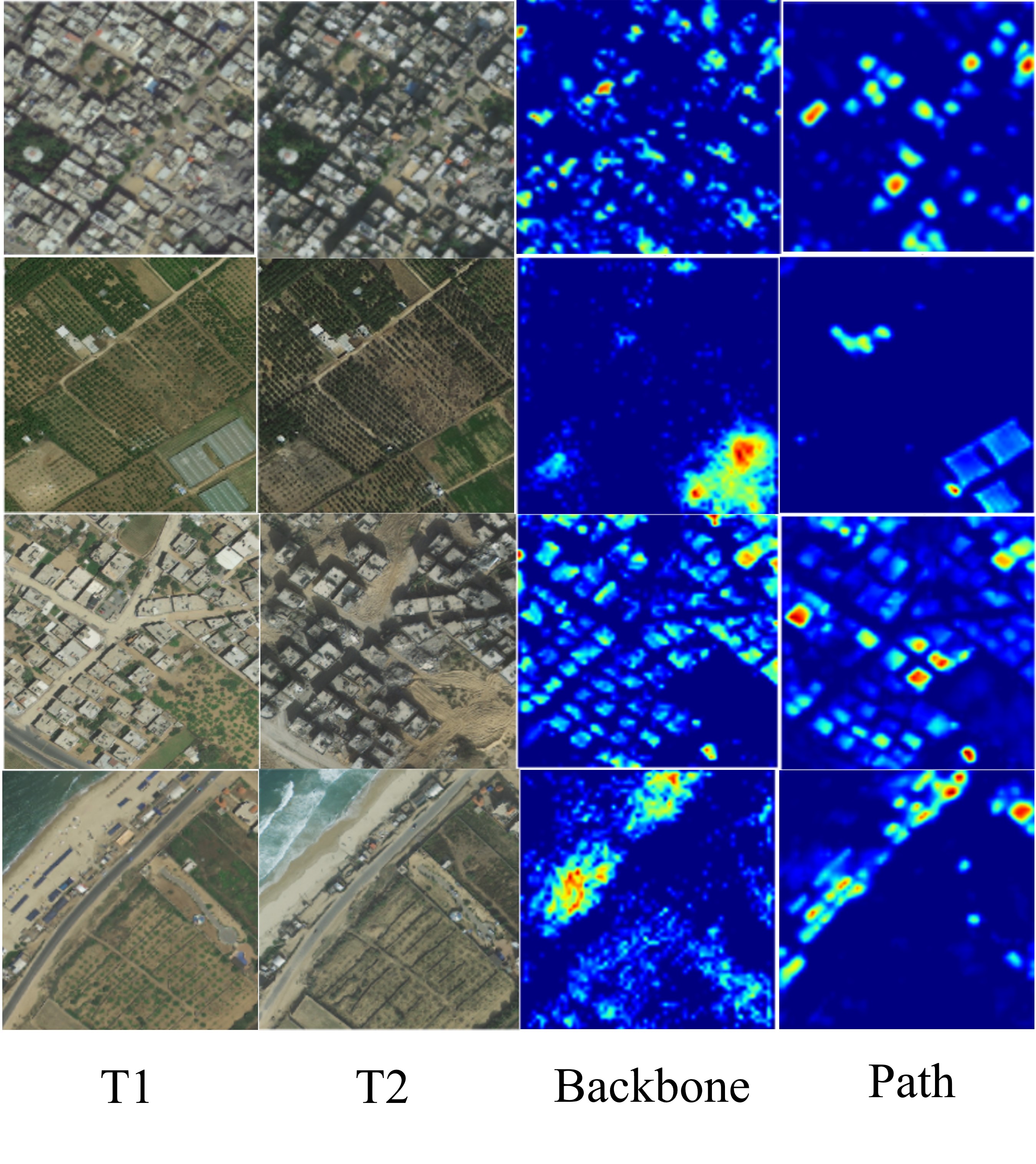}
    \caption{Heatmaps based on the GradCAM visualization of the Gaza-Change dataset. Path represents the features learned by the third pathway, while Backbone represents the features learned by the backbone siamese network.}
    \label{fig:5}
\end{figure*}

\begin{figure*}[htpb]
    \centering
    \includegraphics[width=0.5\linewidth]{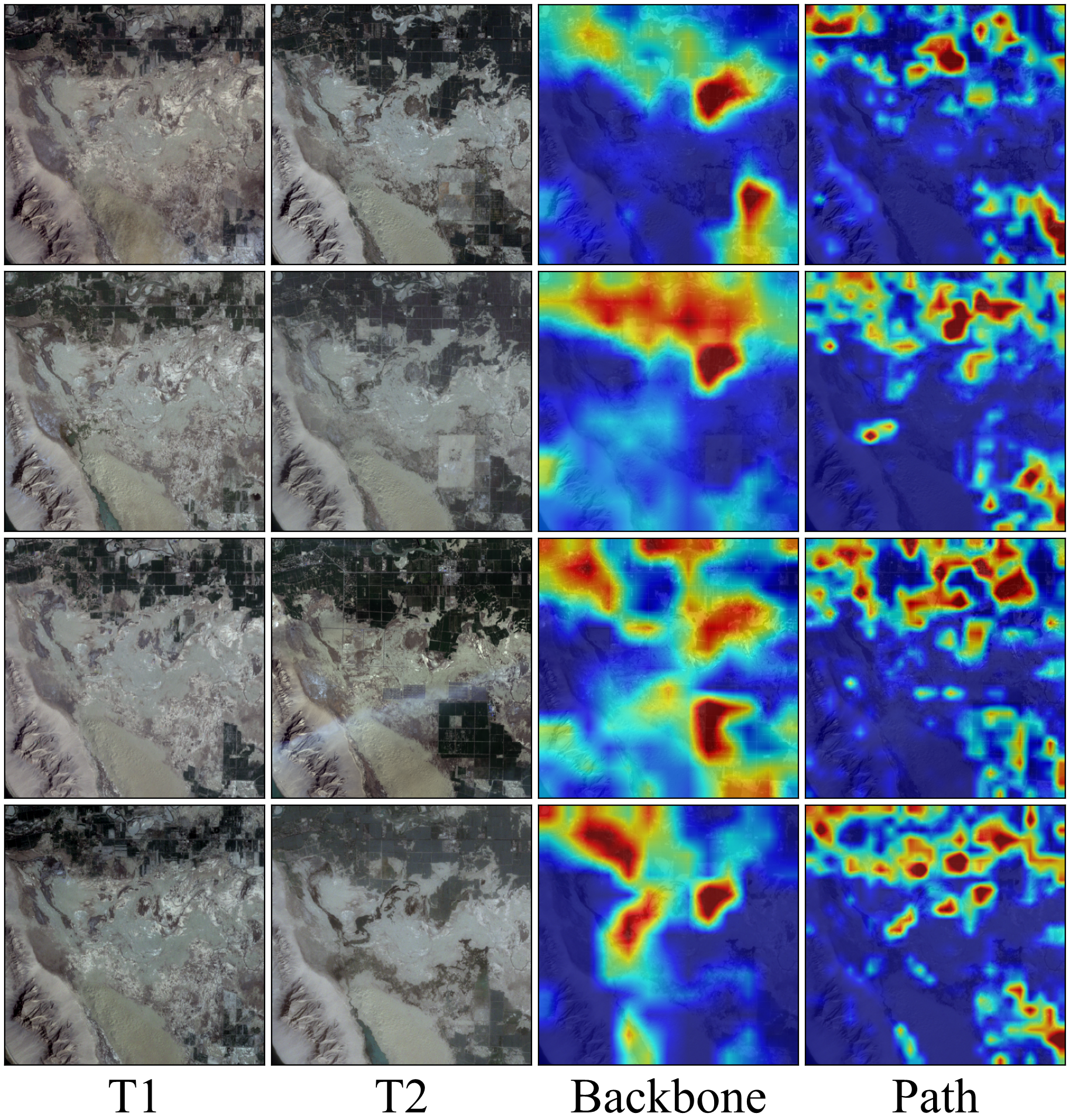}
    \caption{Heatmaps based on the GradCAM visualization of the LandsatSCD dataset. Path represents the features learned by the third pathway, while Backbone represents the features learned by the backbone siamese network.}
    \label{fig:landsatgrad}
\end{figure*}

\begin{figure*}[htpb]
    \centering
    \includegraphics[width=0.5
    \linewidth]{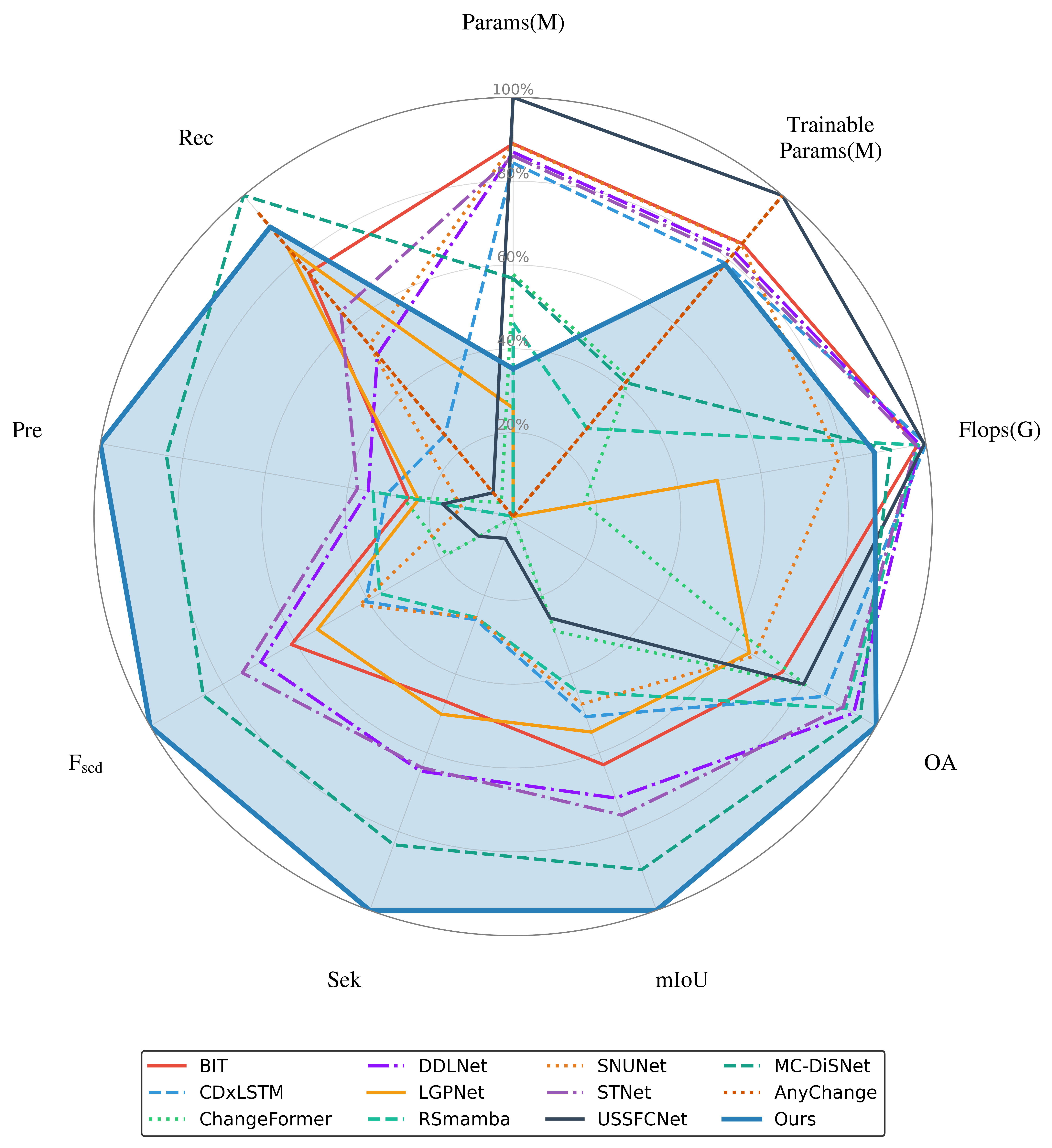}
    \caption{Radar chart of comprehensive performance comparison between the proposed method and existing change detection models.}
    \label{fig:6}
\end{figure*}
As reported in Table~\ref{tab:main}, our proposed Tri-path DINO achieves the best performance across all four evaluation metrics. Specifically, our model attains an OA of 96.45\% and an mIoU of 71.35\%, representing improvements of 0.12\% and 0.71\%, respectively, over the previous best method, MC-DiSNet. This validates the superiority of our model in terms of overall classification accuracy and segmentation consistency across various change categories. More importantly, on the Sek and $F_{scd}$ metrics, which better reflect a model's ability to distinguish between imbalanced change classes, our method achieves significant leads of 13.02\% and 62.58\%, substantially outperforming all other compared methods.

As shown in Figure~\ref{fig:6}, Tri-path DINO achieves the best results on Gaza-change across the four main evaluation metrics, OA, mIoU, SeK, and $F_{\mathrm{scd}}$, demonstrating the effectiveness of complementary feature learning. By freezing the pretrained backbone and introducing lightweight adaptation modules, the model updates 16.38M parameters, accounting for approximately 26.33\% of its total parameters, reflecting its parameter-efficient adaptation. Its computational cost in GFLOPs is also lower than that of comparison methods such as ChangeFormer and LGPNet. These results demonstrate the advantages of the proposed method in detection performance and parameter-efficient adaptation. The associated performance gains and computational costs are further examined in the ``Discussion and Limitations'' subsection.

\begin{table*}[ht]
    \centering
    \caption{
        Comprehensive comparison with state-of-the-art methods on the Gaza-Change dataset. Best results are highlighted in \textbf{bold}.
    }
    \vspace{1mm}
    \label{tab:main}
    \begin{tabular}{c||c|c|c|cccccc}
        \toprule
        \rowcolor{mygray}
        \multicolumn{1}{c||}{}
        & \multicolumn{3}{c|}{Model Size}
        & \multicolumn{6}{c}{Gaza-Change} \\
        
        \rowcolor{mygray}
        Method
        & Params (M)
        & \makecell[c]{Trainable\\Params (M)}
        & FLOPs (G)
        & OA
        & mIoU
        & SeK
        & $F_{\mathrm{scd}}$
        & P
        & R \\
        \hline \hline
        BIT                & 11.89 & 11.89 & 35.38  & 95.74 & 68.81 & 8.91  & 57.85&51.71 & 56.48 \\
        CDxLSTM  & 16.18 & 16.18 & 15.45  & 96.06 & 67.97 & 7.30  & 55.36&54.79 &47.06 \\
        ChangeFormer  & 41.09 & 41.09 & 814.43 & 95.92 & 66.48 & 5.29  & 52.64&52.54&43.15 \\
        DDLNet         & 13.78 & 13.78 & 29.44  & 96.28 & 69.39 & 10.28 & 58.87&57.41&51.73 \\
        LGPNet          & 70.99 & 70.99 & 503.14 & 95.49 & 68.24 & 9.16  & 56.97&50.29&57.91 \\
        RSmamba        & 51.95 & 51.95 & 33.01  & 96.22 & 67.53 & 7.26  & 54.89&56.72&42.34 \\
        SNUNet         & 12.04 & 12.04 & 219.50 & 95.53 & 67.75 & 7.26  & 55.58 &44.13&52.29\\
        STNet          & 14.62 & 14.62 & 38.46  & 96.20 & 69.69 & 10.21 & 59.49&58.97&54.25 \\
        USSFCNet     & 1.52  & 1.52  & 19.49  & 95.90 & 66.25 & 5.70  & 51.57&46.87&43.72 \\
        MC-DiSNet       & 41.99 & 41.99 & 97.98  & 96.33 & 70.64 & 11.73 & 60.80&86.10 & \bf60.98\\
        AnyChange   & 95.22    & 1.48    & 980.69     & 93.70 & 64.48 & 5.27  & 50.42 &36.82 &60.04 \\
        \rowcolor[gray]{0.7} Ours    & 62.21 &16.38 & 135.17     & \textbf{96.45} & \textbf{71.35} & \textbf{13.02} & \textbf{62.58}&\bf95.49 &59.15 \\
        \hline
    \end{tabular}
    \vspace{-4mm}
\end{table*}

\begin{table*}[ht]
    \centering
    \caption{
        Comprehensive comparison with state-of-the-art methods on the SECOND dataset. 
        Best results are highlighted in \textbf{bold}.
    }
    \vspace{1mm}
    \label{tab:main2}
    \begin{tabular}{c||cccccc}
        \toprule
        \rowcolor{mygray}
        \multicolumn{1}{c||}{} & \multicolumn{6}{c}{SECOND} \\
        \rowcolor{mygray}
        \multicolumn{1}{c||}{\multirow{-2}{*}{Method}} 
        & OA & mIoU & Sek & F\textsubscript{scd}& P & R \\
        \hline \hline
        BIT               & \bf86.77 & 70.79 & 19.28 & 54.52&43.74&25.51 \\
        CDxLSTM  & 86.29 & 70.38 & 17.89 & 52.58&37.25&21.95\\
        ChangeFormer  & 86.36 & 70.26 & 18.35 & 53.55 &36.89&23.46 \\
        DDLNet        & 86.47 & 71.12 & 19.06 & 53.78&40.01&25.88 \\
        LGPNet         & 86.57 & 69.67 & 17.49 & 53.05 &45.86&20.86\\
        RSmamba       & 86.51 & 69.64 & 17.63 & 52.71&39.38&23.20 \\
        SNUNet        & 85.82 & 68.61 & 15.29 & 49.69&34.93&15.96 \\
        STNet          & 86.65 & 68.61 & 18.84 & 53.65&37.23&23.79 \\
        USSFCNet      & 85.52 & 68.70 & 14.93 & 49.45 &28.77&13.96\\
        MC-DiSNet       & 85.97 & 71.14 & 19.75 & 53.91&50.31&31.72 \\
        AnyChange    & 65.84 & 51.94 & 9.37  & 36.61&22.63&24.88 \\
        \rowcolor[gray]{0.7} Ours    & 86.33 & \textbf{71.62} & \textbf{20.61} & \textbf{55.44}&\bf55.54&\bf33.46 \\
        \bottomrule
    \end{tabular}
    \vspace{-4mm}
\end{table*}

We compared various methods on the classic SECOND dataset. Table \ref{tab:main2} shows that Tri-path DINO achieves highly competitive performance on this standard benchmark. Notably, our method attains the highest scores in three key metrics: mIoU (71.62\%), Sek (20.61\%), and $F_{scd}$ (55.44\%). Compared to the previous state-of-the-art method, MC-DiSNet, Tri-path DINO outperforms it across all four evaluated metrics, demonstrating the effectiveness of the complementary feature learning strategy. In contrast, the AnyChange method, based on a pre-trained foundational model, performs significantly below expectations due to its difficulty in handling shifts in input distribution, highlighting the necessity of our fine-tuning strategy.

Table~\ref{tab:main2} presents the quantitative comparison of different methods on the LandsatSCD dataset. Tri-path DINO achieves an OA of 95.09\%, an mIoU of 86.43\%, a SeK of 50.95\%, and an $F_{\mathrm{scd}}$ of 85.31\%, obtaining the highest scores among the compared methods on all four metrics. Relative to MC-DiSNet, the improvements are 0.46, 1.04, 2.89, and 1.27 percentage points, respectively. Together with the results on Gaza-change and SECOND, these findings demonstrate the competitive performance of Tri-path DINO across the three datasets and support the effectiveness and applicability of the proposed complementary feature learning framework in different remote sensing scenarios.

Table~\ref{tab:main_landsat} presents the quantitative comparison of different methods on the LandsatSCD dataset. Tri-path DINO achieves an OA of 95.09\%, an mIoU of 86.43\%, a SeK of 50.95\%, and an $F_{\mathrm{scd}}$ of 85.31\%, obtaining the highest scores among the compared methods on all four metrics. Relative to MC-DiSNet, the improvements are 0.46, 1.04, 2.89, and 1.27 percentage points, respectively. Together with the results on Gaza-change and SECOND, these findings demonstrate the competitive performance of Tri-path DINO across the three datasets and support the effectiveness and applicability of the proposed complementary feature learning framework in different remote sensing scenarios.

       \begin{table*}[ht]
    \centering
    \caption{
        Comprehensive comparison with state-of-the-art methods on the LandsatSCD dataset.
        Best results are highlighted in \textbf{bold}.
    }
    \vspace{1mm}
    \label{tab:main_landsat}
    \begin{tabular}{c||cccccc}
        \toprule
        \rowcolor{mygray}
           \multicolumn{1}{c||}{} & \multicolumn{6}{c}{LandsatSCD} \\
        \rowcolor{mygray}
        \multicolumn{1}{c||}{\multirow{-2}{*}{Method}}

        & OA & mIoU & Sek & F\textsubscript{scd} & P & R \\
        \midrule
        BIT          & 93.62 & 82.80 & 41.92 & 80.82 & 85.79 & 76.08 \\
        CDxLSTM      & 92.35 & 79.85 & 34.41 & 76.77 & 81.73 & 64.23 \\
        ChangeFormer & 92.79 & 80.35 & 35.28 & 77.41 & 82.95 & 57.77 \\
        DDLNet       & 93.52 & 82.59 & 41.25 & 80.75 & 86.24 & 74.01 \\
        RSmamba      & 93.32 & 81.31 & 38.02 & 79.11 & 85.25 & 61.76 \\
        SNUNet       & 91.52 & 78.30 & 30.70 & 73.75 & 73.98 & 57.47 \\
        STNet        & 93.62 & 82.86 & 41.91 & 81.04 & 86.19 & 74.63 \\
        USSFCNet     & 91.35 & 77.75 & 29.72 & 73.32 & 73.34 & 53.65 \\
        MC-DiSNet    & 94.63 & 85.39 & 48.06 & 84.04 & 87.57 & 78.95 \\
        AnyChange    & 89.20 & 70.31 & 17.01 & 65.06 & 81.77 & 39.48 \\
        \rowcolor{gray!30} Ours
        & \textbf{95.09} & \textbf{86.43}
        & \textbf{50.95} & \textbf{85.31}
        & \textbf{88.91} & \textbf{80.69} \\
        \bottomrule
    \end{tabular}
    \vspace{-4mm}
\end{table*}

\subsection{Ablation Study}

We conducted systematic ablation experiments on the Gaza-change, SECOND, and LandsatSCD datasets to validate the effectiveness of the proposed third path and the MLHA module. The experimental results are shown in Tables~\ref{tab:module_ablation1}--\ref{tab:module_ablation_landsatscd} for Gaza-change, SECOND, and LandsatSCD, respectively, where Path indicates whether the third path is introduced, and MLHA indicates whether the Multi-Level Hybrid Attention module is used in the decoder. Compared with the baseline, adding MLHA alone improves OA, mIoU, and \(F_{\mathrm{scd}}\) on all three datasets, while adding the third path with MLHA enabled further improves all four metrics. On LandsatSCD, the additional gains in mIoU and \(F_{\mathrm{scd}}\) are 0.31 and 0.93 percentage points, respectively. This demonstrates the important contribution of multi-level attention fusion and complementary feature modeling to improving MCD performance. When both Path and MLHA are enabled (i.e., the complete Tri-path DINO), the model achieves the best performance across all key metrics on all three datasets among these architectural configurations. These results demonstrate the effectiveness of combining the encoder and decoder module designs.

To further examine the individual contributions of the backbone adaptation modules, we conducted component ablation experiments on LandsatSCD. Table~\ref{tab:ablation_adaptation} reports the results of separately removing LoRA or adapters from the complete model while retaining Prompt and the other components. Removing LoRA reduces Mean F1 from 82.72\% to 76.98\%, highlighting its contribution to backbone adaptation. Removing adapters slightly increases Mean F1 to 82.84\%, indicating that the current results do not establish an additional performance benefit from adapters. Here, Mean F1 denotes the average F1 over change categories and is distinct from \(F_{\mathrm{scd}}\).

We also compared different loss configurations on LandsatSCD to evaluate the adopted composite loss, with the results summarized in Table~\ref{tab:ablation_loss}. The Focal, Dice, and Lov\'asz weights of 0.4/0.3/0.3, following [43], achieve the best performance among the evaluated configurations, with a Mean F1 of 84.35\%, a SeK of 50.95\%, an \(F_{\mathrm{scd}}\) of 85.31\%, and an mIoU of 86.43\%. These results support the adopted loss configuration, which combines objectives emphasizing difficult samples, region overlap, and IoU optimization.

\begin{table}[ht]
  \centering
  \caption{Module-level ablation results on the Gaza-change dataset. Best scores are in \textbf{bold}.}
  \label{tab:module_ablation1}
  \begin{tabular}{lclclclclclcl}
\toprule
 Path  & MLHA  & OA & mIoU &  Sek & $F_{scd}$ \\
\hline
  
    &  & 95.63 & 69.66 &  11.79& 59.42 \\
 & \checkmark  & 96.33 & 70.64 & 11.73 &   60.8 \\
  \checkmark & \checkmark &  \textbf{96.45}& \textbf{71.35}  & \textbf{13.02}  & \textbf{62.58} \\

\bottomrule
  \end{tabular}
\end{table}

\begin{table}[ht]
  \centering
  \caption{Module-level ablation results on the SECOND dataset. Best scores are in \textbf{bold}.}
  \label{tab:module_ablation2}
  \begin{tabular}{lclclclclclcl}
\toprule
 Path  & MLHA  & OA & mIoU &  Sek & $F_{scd}$ \\
\hline
  &  & 85.46 & 70.22 & 18.6 &   53.26 \\
    & \checkmark  &85.97& 71.14 &  19.7& 53.91 \\
 
  \checkmark & \checkmark & \textbf{86.33} &\textbf{71.62}&\textbf{20.61}&\textbf{55.44} \\

\bottomrule
  \end{tabular}
\end{table}

\begin{table}[ht]
  \centering
  \caption{Module-level ablation results on the LandsatSCD dataset. Best scores are in \textbf{bold}.}
  \label{tab:module_ablation_landsatscd}
  \begin{tabular}{lccccc}
\toprule
 Path & MLHA & OA & mIoU & SeK & $F_{\mathrm{scd}}$ \\
\hline
  &  & 94.63 & 85.39 & 48.06 & 82.72 \\
  & \checkmark & 94.81 & 86.12 & 49.63 & 84.38 \\
  \checkmark & \checkmark & \textbf{95.09} & \textbf{86.43} & \textbf{50.95} & \textbf{85.31} \\
\bottomrule
  \end{tabular}
\end{table}

\begin{table*}[ht]
  \centering
  \caption{Adaptation-module ablation results on LandsatSCD. Mean F1 denotes the average F1 over change categories and is distinct from \(F_{\mathrm{scd}}\). Best scores are in \textbf{bold}.}
  \label{tab:ablation_adaptation}
    \begin{tabular}{ccccrrr}
        \toprule
        Path & LoRA & Adapter
        & \makecell{Trainable params\\(M)}
        & \makecell{Mean F1\\(\%)}
        & \makecell{$\Delta$F1\\(pp)}
        & \makecell{SeK\\(\%)} \\
        \midrule
        \checkmark & \checkmark & \checkmark
        & 16.38 & \textbf{84.35} & 0.00 & \textbf{50.95} \\
        \checkmark & & \checkmark
        & 15.32 & 76.98 & $-7.37$ & 39.11 \\
        \checkmark & \checkmark &
        & 15.42 & 82.84 & $-1.51$ & 48.51 \\
        \bottomrule
    \end{tabular}
\end{table*}

\begin{table*}[t]
\centering
\small
\setlength{\tabcolsep}{5pt}
\renewcommand{\arraystretch}{1.13}
\caption{Comparison of loss configurations on LandsatSCD. Best scores are highlighted in bold.}
\label{tab:ablation_loss}
\begin{tabular}{@{}cccrrrr@{}}
\toprule
Focal & Dice & Lov\'asz & Mean F1 (\%) & SeK (\%) & \(F_{\mathrm{scd}}\) (\%) & mIoU (\%) \\
\midrule
\textbf{0.4} & \textbf{0.3} & \textbf{0.3} & \textbf{84.35} & \textbf{50.95} & \textbf{85.31} & \textbf{86.43} \\
0.6 & 0.2 & 0.2 & 83.09 & 49.01 & 84.39 & 85.77 \\
0.2 & 0.6 & 0.2 & 83.14 & 48.80 & 84.33 & 85.69 \\
0.2 & 0.2 & 0.6 & 82.32 & 47.47 & 83.85 & 85.23 \\
1.0 & 0.0 & 0.0 & 82.63 & 49.79 & 84.59 & 86.08 \\
0.0 & 1.0 & 0.0 & 79.83 & 47.36 & 83.48 & 85.40 \\
0.0 & 0.0 & 1.0 & 79.82 & 44.48 & 82.39 & 84.19 \\
0.5 & 0.5 & 0.0 & 83.28 & 49.59 & 84.56 & 85.97 \\
0.5 & 0.0 & 0.5 & 82.73 & 48.04 & 83.99 & 85.41 \\
0.0 & 0.5 & 0.5 & 80.19 & 46.12 & 83.11 & 84.86 \\
0.333 & 0.333 & 0.333 & 82.87 & 48.64 & 84.29 & 85.62 \\
\bottomrule
\end{tabular}
\par\smallskip
\begin{minipage}{0.94\linewidth}
\footnotesize

\end{minipage}
\end{table*}

\subsection{Grad-CAM Visualization}
To intuitively validate the complementary roles of each path in Tri-path DINO and elucidate the model's decision-making mechanism, we employ the Grad-CAM method to visualize its regions of focus. As shown in Figures \ref{fig:4}, \ref{fig:5}, and \ref{fig:landsatgrad}, the primary backbone path predominantly activates coarse-grained semantic regions (such as entire building areas). In contrast, the features from the third pathway concentrate on local, previously unattended areas. This clear functional division demonstrates that the two paths inherently capture complementary information: the primary path provides high-level semantic localization, while the third path supplies detailed contextual evidence.

\begin{figure*}[htpb]
    \centering
    \includegraphics[width=0.5\linewidth]{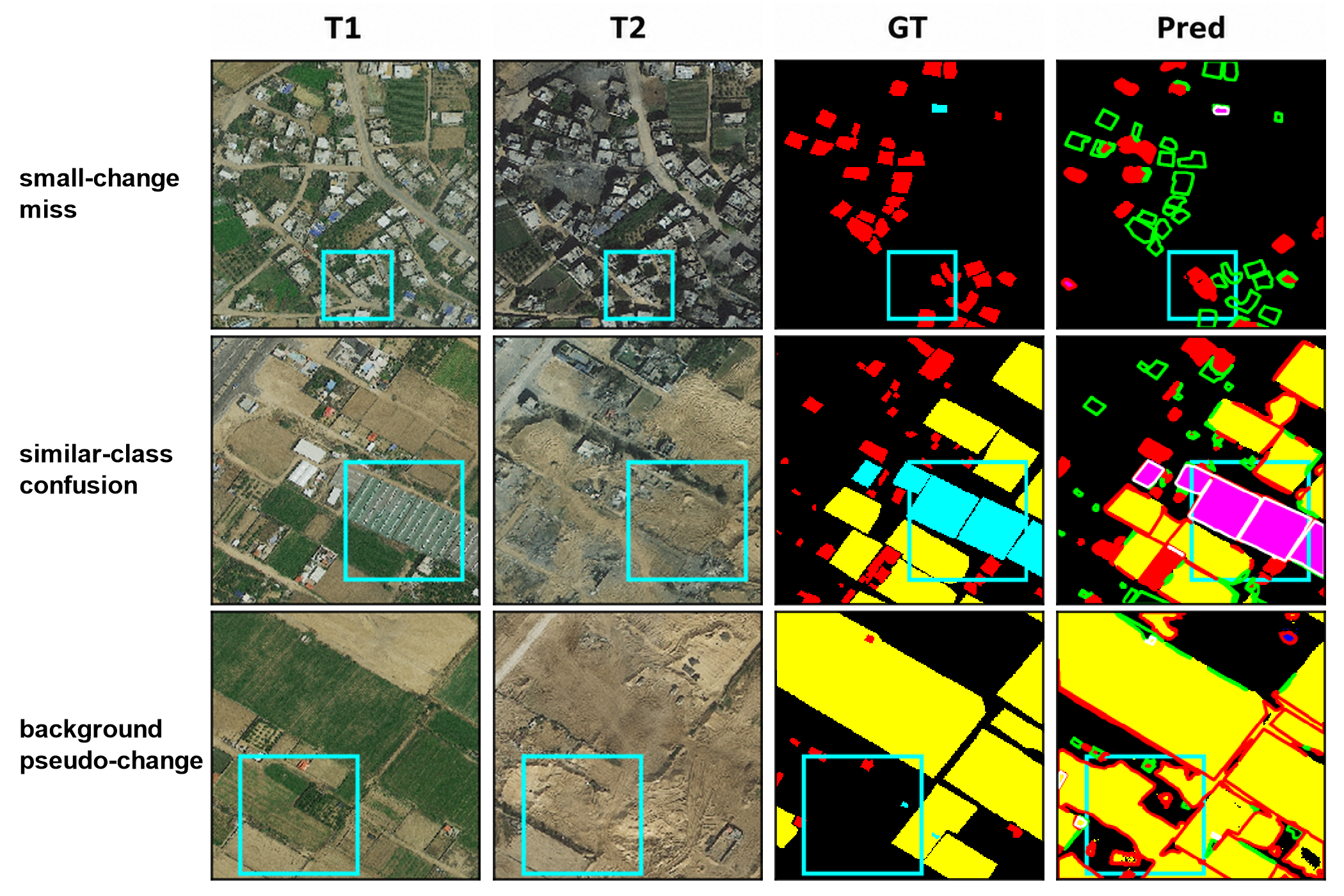}
    \caption{Representative failure cases on the Gaza-Change dataset.
    Columns show T1, T2, ground truth (GT), and prediction (Pred).
    From top to bottom: small-change miss, similar-class confusion,
    and background pseudo-change. Cyan boxes indicate the regions of interest.}
    \label{fig:failure_gaza}
\end{figure*}

\begin{figure*}[htpb]
    \centering
    \includegraphics[width=0.5\linewidth]{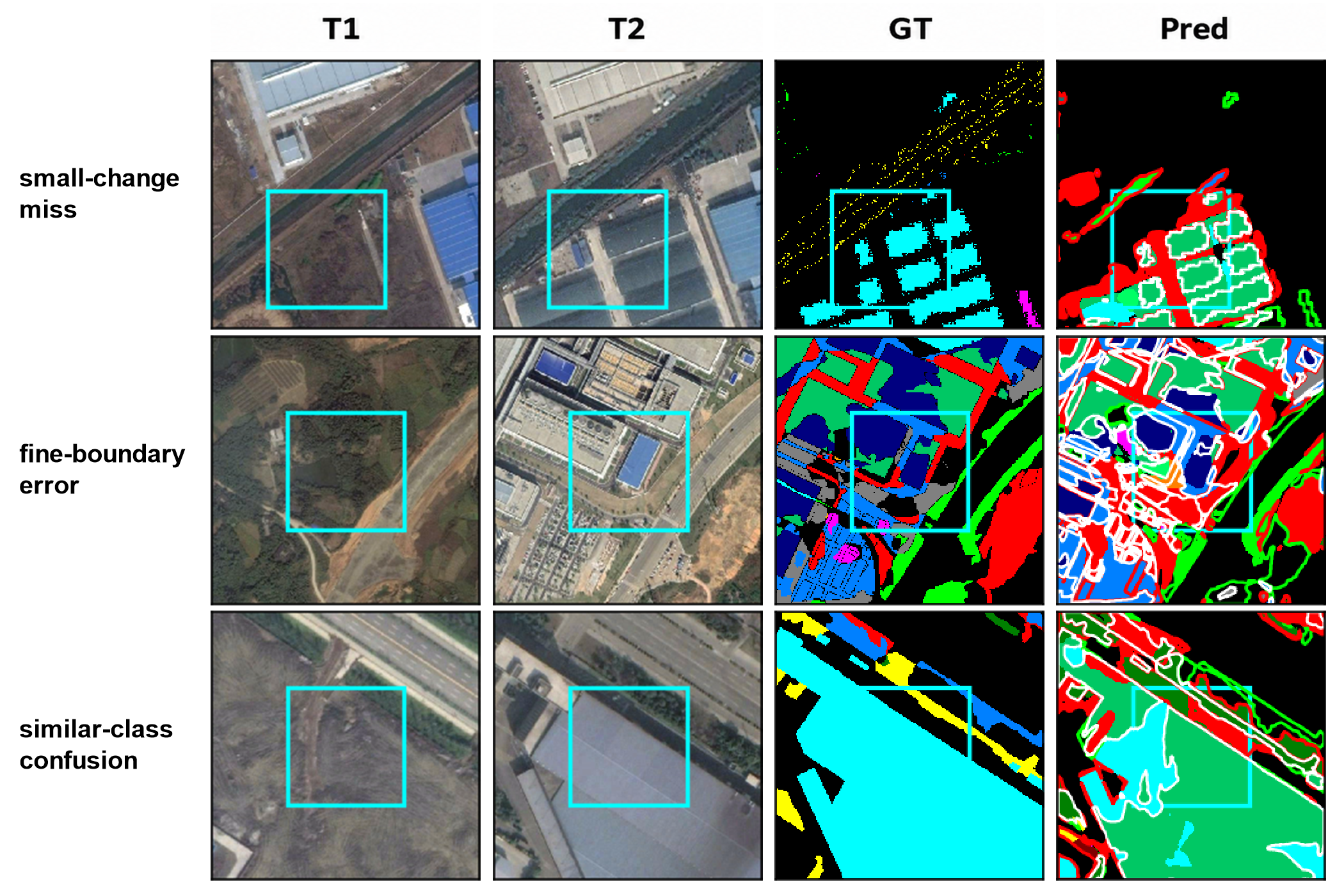}
    \caption{Representative failure cases on the SECOND dataset.
    Columns show T1, T2, ground truth (GT), and prediction (Pred).
    From top to bottom: small-change miss, fine-boundary error,
    and similar-class confusion. Cyan boxes indicate the regions of interest.}
    \label{fig:failure_second}
\end{figure*}

\begin{figure*}[htpb]
    \centering
    \includegraphics[width=0.5\linewidth]{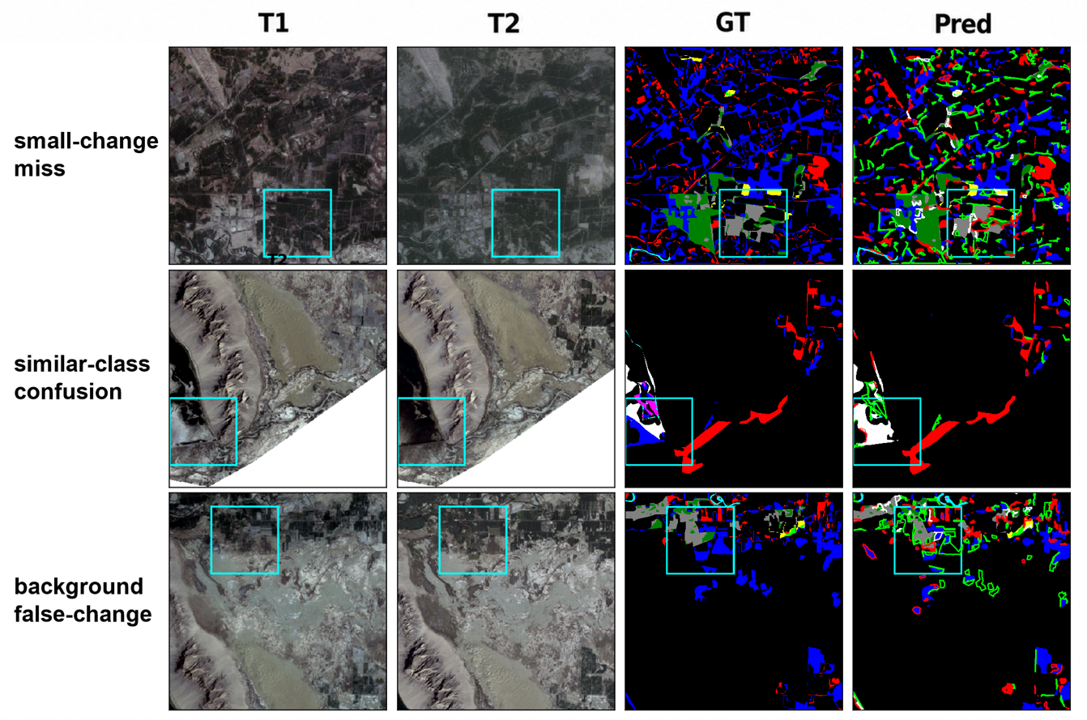}
    \caption{Representative failure cases on the LandsatSCD dataset.
    Columns show T1, T2, ground truth (GT), and prediction (Pred).
    From top to bottom: small-change miss, similar-class confusion,
    and background false-change. Cyan boxes indicate the regions of interest.}
    \label{fig:failure_landsatscd}
\end{figure*}
\subsection{Disscusion}

The strong detection results support the value of complementary feature learning for fine-grained change assessment. Its practical use also depends on the available computational resources. Freezing the backbone limits the parameters updated during training, while the complete backbone still participates in the forward pass and feature aggregation and fusion require additional computation. Thus, parameter-efficient adaptation and overall computational efficiency should be considered separately. The detection gains are useful for applications requiring detailed damage assessment, while reducing computational overhead remains an important direction for deployment under limited resources.

The failure cases in Figs.~\ref{fig:failure_gaza}, ~\ref{fig:failure_second} and~\ref{fig:failure_landsatscd} show incomplete detection of small or fragmented changes, inaccurate boundaries, category confusion, and false positives in unchanged regions. On Gaza-change, regions annotated as new greenhouses are predicted as greenhouse damage, illustrating that locating changes does not ensure correct categorization. The SECOND and LandsatSCD examples also show loss of fine structures and discrepancies in the predicted contours. These observations indicate remaining limitations in spatial localization and semantic discrimination. Future work will focus on small-target detection, boundary refinement, and stronger discrimination between change categories.

\section{Conclusion}
This paper proposes a novel network architecture for multi-class change detection in high-resolution remote sensing imagery, which can be applied to remote sensing monitoring scenarios. To address the challenges of MCD, where target regions are extremely small and inter-class feature differences are subtle, we designed a context-complementary feature learning architecture, utilizing a pre-trained DINOv3 model as the feature extractor. The backbone network employs a CNN for feature extraction to enhance the learning of coarse-grained features. Intermediate multi-layer features are fed into a Transformer feature extractor to achieve fine-grained feature learning. A Multi-Level Hybrid Attention (MLHA) module is incorporated into the decoder to improve adaptive multi-scale feature refinement. Extensive experiments on the real-world challenging Gaza-change damage assessment dataset and the standard SECOND benchmark demonstrate the effectiveness of the proposed method. Our work confirms that the MCD framework offers a practical and annotation-efficient balance between binary change detection and full semantic change detection, especially suitable for time-sensitive, fine-grained applications such as post-disaster damage assessment. Although Tri-path DINO has shown strong performance, future work may extend it to multi-temporal change analysis and explore more efficient designs for real-time processing.

\printcredits

\bibliographystyle{cas-model2-names}

\bibliography{ijcai26.bib}

\section*{Declaration of Competing Interest}
The authors declare that they have no conflict of interest in this paper.

\end{document}